# Tangent-based manifold approximation with locally linear models

Sofia Karygianni, *Student Member, IEEE,* and Pascal Frossard, *Senior Member, IEEE,*

**Abstract**

In this paper, we consider the problem of manifold approximation with affine subspaces. Our objective is to discover a set of low dimensional affine subspaces that represents manifold data accurately while preserving the manifold's structure. For this purpose, we employ a greedy technique that partitions manifold samples into groups that can be each approximated by a low dimensional subspace. We start by considering each manifold sample as a different group and we use the difference of tangents to determine appropriate group mergings. We repeat this procedure until we reach the desired number of sample groups. The best low dimensional affine subspaces corresponding to the final groups constitute our approximate manifold representation. Our experiments verify the effectiveness of the proposed scheme and show its superior performance compared to state-of-the-art methods for manifold approximation.

## I. INTRODUCTION

The curse of dimensionality is one of the most fundamental issues that researchers, across various data processing disciplines, have to face. High dimensional data that is difficult to even store or transmit, huge parametric spaces that are challenging to exploit and complex models that are difficult to learn and prone to over-fitting, are some simple evidences of the dimensionality problem. However, it is not rare that the data presents some underlying structure, which can lead to more efficient data representation and analysis if modeled properly.

S. Karygianni and P. Frossard are with Ecole Polytechnique Fédérale de Lausanne (EPFL), Signal Processing Laboratory-LTS4, CH-1015, Lausanne, Switzerland (e-mail:{sofia.karygianni, pascal.frossard}@epfl.ch).

In many cases, the underlying structure of the signals of a given family can be described adequately by a manifold model that has a smaller dimensionality than the signal space. Prominent examples are signals that are related by transformations, like images captured under different viewpoints in a 3D scene, or signals that represent different observations of the same physical phenomenon like the EEG and the ECG. Manifolds have already been adopted with success in many different applications like transformation-invariant classification, recognition and dimensionality reduction [1], [2], [3].

In general, manifolds are topological spaces that locally resemble a Euclidean space. Therefore, although as a whole they might be extremely complicated structures, locally, i.e., in the neighborhood of a point, they have the same characteristics as the usual Euclidean space. In this work, we are going to consider $d$-dimensional, differentiable manifolds that are embedded into a higher dimensional Euclidean space, $\mathbb{R}^N, N >> d$. Intuitively, one can think of a $d$-dimensional manifold embedded into $\mathbb{R}^N$ as the generalization of a surface in $N$ dimensions: it is a set of points that locally seem to live in $\mathbb{R}^d$ but that macroscopically synthesize a structure living into $\mathbb{R}^N$. For example, a sphere in $\mathbb{R}^3$ and a circle in $\mathbb{R}^2$ are both manifolds of dimension 2 and 1 respectively. Although manifolds are appealing for effective data representation, their unknown and usually strongly non-linear structure makes their manipulation quite challenging. There are cases where an analytical model can represent the manifold, like a model built on linear combinations of atoms coming from a predefined dictionary [4]. However, an analytical model is unfortunately not always available. A workaround consists in trying to infer a global, data-driven parametrization scheme for the manifold by mapping the manifold data from the original space to a low-dimensional parametric space. The problem of unveiling such a parametrization is called manifold learning.

Usually, it is hard to discover a universal manifold representation that is always accurate as it means that all the non-linearities of the manifold are well represented by only one mapping function. Therefore, instead of using just one global scheme, it is often preferable to employ a set of simpler structures to approximate the manifold's geometry. This can be done in the original space where the manifold lives. The objective of the approximation is to create a manifold model that is as simple as possible while preserving the most crucial characteristic of a manifold: its shape. An example of such an approximation for an 1D manifold is shown in Figure 1a, where a set of lines represents the spiral shape.

In this paper, we approximate generic manifolds with the simplest possible models, the affine subspaces (flats). Such a choice is motivated by the locally linear character of a manifold as well as the simplicity and efficiency of flats for performing local computations like projections. Our objective is to compute a set of low dimensional flats that represents the data as accurately as possible and at the same time preserves the geometry of the underlying manifold. We formulate the manifold approximation problem as a constrained clustering problem for manifold samples. The constraints are related to the underlying geometry of the manifold, which is expressed by the neighborhood graph of the data samples. We borrow elements of the constrained clustering theory to motivate the use of a greedy approximation scheme. For choosing our optimization function, we relate the capability of a set of points to be represented by a flat, with the variance of the tangents at these points. Then, we use the difference of tangents to uncover groups of points that comply with the low dimensionality of flats. The partitioning is done in a bottom-up manner where each manifold sample is considered as a different group at the beginning. Groups are then iteratively merged until their number reduces to the desired value. We have tested our algorithm on both synthetic and real data where it gives a superior performance compared to state-of-the-art manifold approximation techniques.

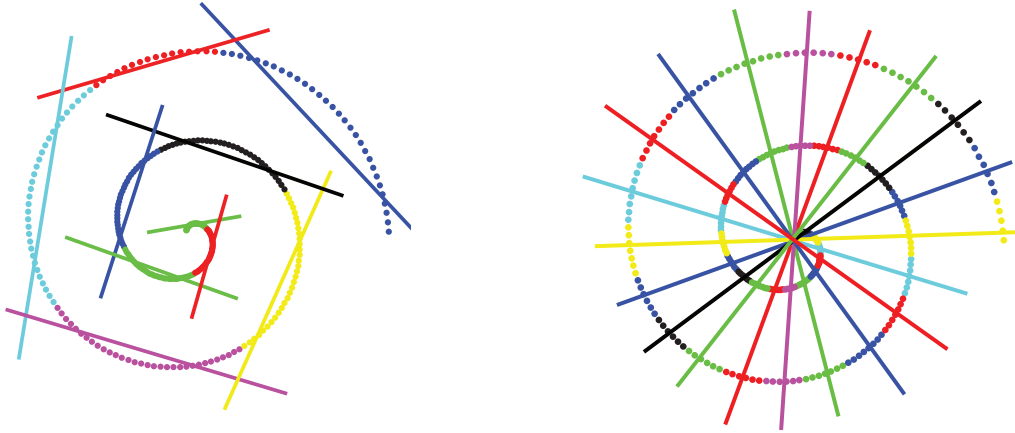

(a) Good manifold approximation example

(b) Bad manifold approximation example

Fig. 1: Manifold approximation illustration. On the left, we have an example of a valid approximation by lines of a 1D manifold embedded into $\mathbb{R}^2$. The different colors represent the different groups of samples, each approximated by a line. On the right, we have an example where the approximation does not align well with the manifold structure, as a result of the median k-flats algorithm [5].

4The rest of the paper is organized as follows. In Section II we discuss the related work in manifold approximation and other relevant fields like manifold learning and hybrid linear modeling. In Section III, we motivate the use of a greedy approximation strategy with concepts from constrained clustering theory and we present our problem formulation. We present in Section IV, our approximation algorithm in detail. In Section V, we describe the experimental setup and the results of our experiments. Finally, in Section VI, we provide concluding remarks.

## II. RELATED WORK

Data representation with affine models has received quite some attention lately. Relative approaches usually fall under the name of either subspace clustering or hybrid linear modeling. Their objective is to find a set of affine models explaining the different data sources, i.e., to cluster the data into groups so that each group can be represented well by a low-dimensional affine space. A common approach is to use an iterative scheme to alternate between steps of data segmentation and subspace estimation aiming at either minimizing the sum of reconstruction errors [5], [6] or maximizing the likelihood of the data under a probabilistic model, like probabilistic PCA [7]. Alternatively, different kinds of algebro-geometric approaches have also been proposed. An interesting formulation has been presented in [8], where the problem of subspace clustering is transformed into a problem of fitting and manipulating polynomials. Moreover, in [9], [10], the spectral analysis of an appropriately defined similarity matrix over the data is used to uncover the underlying low dimensional structures as well as the partition that favors them. Recently, in [11], the use of spectral analysis is combined with a multiscale analysis of the rate of growth of the local neighborhoods' eigenvalues, so that, apart from the appropriate clustering, the model parameters, number and dimensionality of the subspaces, are simultaneously recovered from the data. While they are quite successful at times, the above methods apply mainly to cases where data is generated from different low dimensional subspaces that do not necessarily form a manifold. And as such, they uncover a set of linear spaces that do not necessarily comply with the manifold structure, such as the set of lines shown in Figure 1b.

As far as manifold-driven data is concerned, there is a great variety of works in the so called field of manifold learning and dimensionality reduction. The goal of manifold learning is to devise a low dimensional, global parametrization for data sets that lie on high dimensional non-linear manifolds, while preserving some properties of the underlying manifold. Two pioneer works in the field are the Isomap [1] and the LLE algorithms [12]. In Isomap, the parametrization is uncovered in a way that preserves the geodesic distances between the points while in LLE the focus is on preserving the local linear properties

November 9, 2012 DRAFT

of neighborhoods. Other well known approaches that aim at preserving local properties of the points' neighborhoods are provided by the Laplacian Eigenmaps (LE) [13] and the Hessian Eigenmaps (HLLE) [14]. Recently, these methods have been extended to points lying on Riemmanian manifolds as opposed to Euclidean spaces [2]. This opens the range of possible applications for manifold-based representations. A detailed list of the most popular algorithms for manifold learning can be found in [15] and [16], along with interesting comments on their relative strengths and weaknesses.

In manifold approximation the goal is however, to represent the manifold structure in the original space. The ultimate target is not a global parametrization, but rather a set of local, affine subspaces that could approximate the original geometry accurately. Although the locally linear nature of manifolds has been used as a tool for learning a global parametrization by aligning or combining local probabilistic data models (e.g., [17] , [18]), only a few works so far have tried to create a model of the manifold in the original space while preserving its structural properties. Two such examples are the works of Wang and Chen [3] and Fan and Yeung [19]. In [3], the authors introduce the Hierarchical Divisive Clustering (HDC) algorithm, which is a method for hierarchically partitioning the data by dividing highly non-linear clusters. As a linearity measure, it uses the deviation between the euclidean and geodesic distances. In [19], the clustering is performed in a bottom-up manner, named Hierarchical Agglomerative Clustering (HAC), where again the geodesic distances are used to express the underlying manifold structure.

In our work, we have chosen a bottom-up approach but we use a different linearity measure, namely the variance of the tangent spaces. As it will be shown in the next section, this measure emerges naturally from the definition of the local properties of a manifold while both linearity measures in [19] and [3] are more simplistic. In fact, the importance of the tangent spaces for manifold related tasks has been recognized by many researchers. In [20] the authors use the tangent spaces to infer valid parametrizations of a manifold, and recently, in [21] the authors focus on the reliable estimation of the tangent spaces from the data. They also incorporate the tangent distance into a variation of a k-means algorithm to classify samples into linear groups, which is another piece of evidence that tangent distances can be used for identifying linear regions on manifolds. Our approach however specifically addresses the problem of linear manifold approximation as it is fuses nicely the tangent distances with the theory of constrained clustering into a simple, and yet effective clustering algorithm.

## III. MANIFOLD APPROXIMATION PROBLEM

*A. Framework*

We consider the problem of approximating a $d$-dimensional manifold $\mathcal{M}$, embedded into $\mathbb{R}^N$, with a set of $d$-dimensional affine subspaces, which we call flats. The dimension $d$ is an external parameter in our problem; in practice it is either specific to the application at hand or estimated a priori from the data. The manifold is represented by the set of samples $\mathcal{X} = \{x_k \in \mathbb{R}^N, k \in [1,m]\}$ and the undirected and symmetric neighborhood graph $G_\mathcal{X} = G(\mathcal{X}, E)$, which represents the manifold's geometry by connecting neighbor samples on the manifold. There exist various ways to construct $E$ when it is not given a priori. We have chosen to use the k-nearest neighbor approach, i.e., we connect each sample in $\mathcal{X}$ with its k-nearest neighbors. Our objective is then to uncover a partition of $\mathcal{X}$ into $\mathcal{L}$ clusters, $\mathbf{C}_\mathcal{L}(\mathcal{X}) = \{C_i, i \in [1,\mathcal{L}]\}$, so that each cluster can be represented well by a $d$-dimensional flat that respects the underlying geometry of the manifold. In order for $\mathbf{C}_\mathcal{L}(\mathcal{X})$ to be a valid partition of $\mathcal{X}$, the involved clusters should not overlap and they should cover the whole set $\mathcal{X}$, i.e., $C_j \cap C_i = \emptyset$, $\forall\, i \neq j$ and $\cup_{i=1}^{\mathcal{L}} C_i = \mathcal{X}$.

There are many different ways to partition a set into $L$ clusters. However, in our case not all possible partitions of $\mathcal{X}$ are valid since we are interested only in partitions that respect the underlying geometry of the manifold. In particular, we consider the partitions whose clusters spread different regions of the manifold to be invalid. Although these clusters can be approximated well with flats, the resulting flats do not comply with the local manifold structure. Such a bad partitioning example is illustrated in Figure 1b. In order to check the compliance of a partition $\mathbf{C}_\mathcal{L}(\mathcal{X})$ with the manifold's shape we can use the graph $G_\mathcal{X}$. Then, a sufficient condition for a partition to be valid is to have clusters with connected subgraphs. To be more specific, each cluster's subgraph is defined as $G_{C_i} = G_\mathcal{X}(C_i, E_i)$ where $E_i = \{a_{ij} \in E : x_i, x_j \in C_i\}$ is the set of edges in $E$ with both endpoints in $C_i$. Then, the subgraph $G_{C_i}$ is connected if every pair of nodes in $C_i$ is connected with a path in $E_i$. The set of all partitions that fulfill this condition is called the *feasible set of order* $\mathcal{L}$ and denoted by $\mathbf{\Phi}_\mathcal{L}(\mathcal{X})$. The corresponding feasibility predicate, $\Phi_\mathcal{X}(\mathbf{C}_\mathcal{L}) \equiv \mathbf{C}_\mathcal{L} \in \mathbf{\Phi}_\mathcal{L}(\mathcal{X})$, is then defined as:

$$\Phi_\mathcal{X}(\mathbf{C}_\mathcal{L}) = \bigwedge_{C_i \in \mathbf{C}_\mathcal{L}} \phi(C_i), \text{ where } \phi(C_i) = \begin{cases} \text{true}, & \text{if } G_{C_i} \text{ is connected} \\ \text{false}, & \text{if } G_{C_i} \text{ is not connected,} \end{cases} \quad (1)$$

where the symbol $\wedge$ stands for logical addition.

Finally, we define the fusibility predicate $\psi(C_i, C_j)$ that expresses the possibility of fusing clusters $C_i$ and $C_j$. It is closely related with the feasibility predicate $\phi$ of Eq. (1) by the following property of

binary heredity:

$$\text{if } C_i, C_j \neq \emptyset, \ C_i \cap C_j = \emptyset, \ \phi(C_i) \wedge \phi(C_j) \text{ and } \psi(C_i, C_j), \text{ then } \phi(C_i, C_j) \qquad (2)$$

This property means that the fusion of two good and related clusters should give a good cluster. In our case, the feasibility predicate is related to the connectivity of the clusters' graphs $G_C$. Therefore, we will allow clusters $C_i$ and $C_j$ to be fused only if the graph corresponding to their union, $G_{C_i \cup C_j}$, is connected. A sufficient condition for that is the existence of an edge between any sample in $C_i$ and any sample in $C_j$.

## B. Tangent-based clustering

Several data partitions are feasible, but we are interested in partitions that effectively capture the manifold's local geometry. In order to evaluate the 'quality' of a feasible partition $\mathbf{C}$, we first need a criterion function $P$ that is non-negative, distributive over the clusters in $\mathbf{C}$ and zero for the case of single-element clusters, i.e.,

$$P(\mathbf{C}) = \sum_{C_i \in \mathbf{C}} p(C_i) \text{ with } p(C_i) \geq 0 \text{ and } p(\{x\}) = 0, \ \forall x \in \mathcal{X}. \qquad (3)$$

The function $p(C_i)$, which represents the distribution of $P$ over the clusters in a partition, is non-negative for all clusters and zero for single-element clusters. Moreover, our goal is to uncover clusters that can be well-represented by $d$-dimensional flats. Therefore, the function $p$ should be measuring the distance between a linear $d$-dimensional space and the manifold points in the corresponding cluster.

According to the definition in [22], a set $\mathcal{M} \subseteq \mathbb{R}^N$ is a $d$-dimensional differentiable manifold iff $\forall x \in \mathcal{M}$ there exist open sets $V \in \mathbb{R}^N$ with $x \in V$ and $W \in \mathbb{R}^d$ as well as a one-to-one, differentiable function $f : W \to \mathbb{R}^N$ with continuous inverse such that

$$f(W) = \mathcal{M} \cap V$$

$$f'(y) = Df(y), \text{ the Jacobian matrix of } f, \text{ has rank } d, \forall y \in W$$

The function $f$ is called a coordinate system at $x$. Assuming that $f(a) = x$, the $d$-rank Jacobian matrix $Df(x)$ and the corresponding linear transformation $f_* : \mathbb{R}_a^d \to \mathbb{R}_x^N$ define a $d$-dimensional subspace of $\mathbb{R}_x^N$, which is the tangent space of $\mathcal{M}$ at $x$ denoted $M_x$.

For some choices of $x, V$ and $W$, the function $f$ might be linear. Then, $Df(a) = Df(b), \forall a, b \in W$, which means that the tangent spaces of all points $x \in \mathcal{M} \cap V$, seen as subspaces of $\mathbb{R}_N$ coincide when they are transferred to the same origin point. These regions can be perfectly represented by flats.

Therefore, the goal of our algorithm is to cluster together the samples that come from such regions on the manifold.

Since our target regions are characterized by low variability of the tangent spaces, it seems appropriate to use a variance-based criterion function $p(C_i)$ that measures the variance of the tangents of the samples in a cluster $C_i$, i.e.,

$$p(C_i) = \sum_{x \in C_i} D_T^2(M_{C_i}, M_x) \qquad (4)$$

where $M_{C_i}$ is the mean tangent over the tangents of the samples in $C_i$ and $D_T$ is a suitably chosen distance measure for the tangents. Instead of working with a set of $d$-dimensional subspaces that are positioned at point $x$, it is more convenient to translate all of them to the origin of $\mathbb{R}^N$. For the rest of the paper, $M_x$ refers to the tangent space of $x$ translated to the origin of $\mathbb{R}^N$.

We give now more details on the computation of the distance $D_T$. The space of all $d$-dimensional linear subspaces of $\mathbb{R}^N$ is called the Grassmann manifold, denoted as $G_{N,d}$ [23]. Each member of the $G_{N,d}$ can be represented by any of its bases. In our case, $M_x$ is actually described by such an orthonormal basis. A unique $N \times N$ projection matrix $P = BB^T$, that is idempotent and of rank $d$, corresponds to each $d$-dimensional subspace of $\mathbb{R}^N$ with basis $B$. The set of all orthogonal projections matrices of rank $d$ is called $P_{d,N-d}$ and is a manifold equivalent to the Grassmann manifold. However, $P_{d,N-d}$ is also embedded in $\mathbb{R}^{N \times N}$. Since $\mathbb{R}^{N \times N}$ is a vector space, we can define the distance between two tangents $M_x$ and $M_y$ as:

$$D_T(M_x, M_y) = \frac{1}{\sqrt{2}} ||P_x - P_y||_F = \frac{1}{\sqrt{2}} ||M_x M_x^T - M_y M_y^T||_F = \left[ d - tr(M_x^T M_y M_y^T M_x) \right]^{1/2} \qquad (5)$$

The distance $D_T(M_x, M_y)$ is called the projection metric, because it results from the projection of $G_{N,d}$ into $P_{d,N-d}$ [24].

Finally, we can use the equivalence between $G_{N,d}$ and $P_{d,N-d}$ to derive a computation procedure for the mean tangent of a cluster $C_i$, given in Eq. (4). A common definition of the mean or center of a set $C$ of points in the metric space $S$ has been given by Karcher in [25] as the element $m_C \in S$ that minimizes the sum of square distances $\mathcal{D}$ to the points $x$ in the set, i.e.,

$$m_C = \arg\min_{s \in S} \sum_{x \in C} \mathcal{D}^2(x, s) \qquad (6)$$

In our case, we are given a cluster $C_i$, where each sample $x \in C_i$ has a tangent space $M_x$ and a corresponding projection matrix $P_x = M_x M_x^T$. We need to compute the mean tangent $M_{C_i}$, with corresponding projection matrix $P_{C_i}$. Using the projection distance introduced in Eq. (5), Eq. (6) translates

into:

$$P_{C_i} = \underset{A \in P_{d,N-d}}{\arg\min} \sum_{x \in C_i} \frac{1}{2}||P_x - A||_F^2 \qquad (7)$$

Since $P_{d,N-d}$ is embedded into $R_{N \times N}$ there is a matrix, $\tilde{P}_{C_i} = \frac{1}{|C_i|} \sum_{x \in C_i} P_x$, that minimizes Eq. (7) in $R^{N \times N}$. However, $\tilde{P}_{C_i}$ does not have to belong to $P_{d,N-d}$. Therefore we need to project it back to $P_{d,N-d}$, i.e., we need to find the matrix that minimizes

$$P_{C_i} = \underset{P \in P_{d,N-d}}{\arg\min} ||P - \tilde{P}_{C_i}||_F \qquad (8)$$

Using the Eckart-Young theorem on low rank approximation of matrices under the Frobenius norm [26], we can find that the matrix the solution of Eq. (8) is $P_{C_i} = UU^T$, where $U$ is the matrix of eigenvectors in the $d$-rank singular value decomposition of $\tilde{P}_{C_i}$, i.e., $\tilde{P}_{C_i} = US_dU^T$. The corresponding subspace on the Grassmann manifold is thus the one spanned by $U$. Therefore, $M_{C_i} = U$.

This procedure for computing the mean of a set on the Grassmann manifold is often referred to as an extrinsic mean computation procedure [27], as it uses the equivalence between $G_{N,d}$ and $P_{d,N-d}$ to perform the mean computation into $R^{N \times N}$ instead of computing the mean into the original space.

Equipped with the above developments, we can now formalize our manifold approximation objective as finding the feasible partition $\mathbf{C}_\mathcal{L}^*(\mathcal{X})$ that minimizes $P$, i.e.,

$$\mathbf{C}_\mathcal{L}^*(\mathcal{X}) = \underset{\mathbf{C} \in \mathbf{\Phi}_\mathcal{L}(\mathcal{X})}{\operatorname{argmin}} P(C) = \underset{\mathbf{C} \in \mathbf{\Phi}_\mathcal{L}(\mathcal{X})}{\operatorname{argmin}} \sum_{C_i \in \mathbf{C}} p(C_i) \qquad (9)$$

By substituting the exact form of the criterion function (4) in (9) we get the following constrained clustering problem:

$$\mathbf{C}_\mathcal{L}^*(\mathcal{X}) = \underset{\mathbf{C} \in \mathbf{\Phi}_\mathcal{L}}{\operatorname{argmin}} \sum_{C_i \in \mathbf{C}} \sum_{x \in C_i} D_T^2(M_{C_i}, M_x) \qquad (10)$$

where $\mathbf{\Phi}_\mathcal{L}$ is defined in (1), $D_T$ is a distance measure on the Grassman manifold and $M_{C_i}$ is the mean tangent of cluster $C_i$. From [28], the constrained clustering problems with the form of Eq. (9) can also be expressed in the form of the generalized Jensen equality [29]:

$$\mathbf{C}_\mathcal{L}^*(\mathcal{X}) = \begin{cases} \{\mathcal{X}\}, & \mathcal{L} = 1 \\ \mathbf{C}_{\mathcal{L}-1}^*(\mathcal{X} \setminus C^*) \cup \{C^*\}, & \mathcal{L} > 1 \end{cases} \qquad (11)$$

where

$$C^* = \underset{\substack{\emptyset \subset C \subset \mathcal{X} \\ \exists \mathbf{C} \in \mathbf{\Phi}_{\mathcal{L}-1}(\mathcal{X} \setminus C): \mathbf{C} \cup \{C\} \in \mathbf{\Phi}_\mathcal{L}(\mathcal{X})}}{\operatorname{argmin}} (P^*(\mathcal{X} \setminus C) + p(C)) \qquad (12)$$

The symbol $\setminus$ stands for set subtraction and $\cup$ for set addition. This is a dynamic programming equation that may lead to polynomial time solutions under certain constraints and characteristics of the clustering

November 9, 2012 DRAFT

problem [30]. However, in the general case, this approach gives rise to algorithms that have exponential time complexity.

An alternative way of solving the above problem is presented in [31]. It allows for more efficient, but less accurate, algorithms. It is a top-down procedure, where the best clustering $\mathbf{C}_{\mathcal{L}}(\mathcal{X})$ is expressed in terms of the clusterings in $\mathbf{\Phi}_{\mathcal{L}+1}$ instead of $\mathbf{\Phi}_{\mathcal{L}-1}$ as in Eq. (11). We opt for such a top-down approach for solving the problem in Eq. (9).

In such a top-down approach, we however need a measure for comparing clusters and deciding on proper merging choices. Thus, we define the dissimilarity measure $d : (C_i, C_j) \to \mathbb{R}_0^+$ as the difference in the criterion function before and after the merging of two clusters, i.e.,

$$d(C_i, C_j) = p(C_i \cup C_j) - p(C_i) - p(C_j), \tag{13}$$

assuming that the merging of any two good clusters gives always rise to a cluster with a higher score in terms of the criterion function. Under some mild assumptions on the relations among $P, d$ and $\Phi$ [31], we can now rewrite Eq. (9) as

$$\mathbf{C}_{\mathcal{L}}^*(\mathcal{X}) = \left(\mathbf{C}'_{\mathcal{L}+1}(\mathcal{X}) \setminus \{C'_i, C'_j\}\right) \cup \{C'_i \cup C'_j\} \tag{14}$$

where

$$(\mathbf{C}'_{\mathcal{L}+1}(\mathcal{X}), C'_i, C'_j) = \underset{\substack{C_i, C_j \in \mathbf{C} \\ \mathbf{C} \in \mathbf{\Phi}_{\mathcal{L}+1} \\ \psi(C_i, C_j) \text{ is true}}}{\operatorname{argmin}} \; (P(\mathbf{C}) + d(C_i, C_j)))$$

This equation just says that, in order to find the best partition $\mathbf{C}_{\mathcal{L}}^*(\mathcal{X})$ we need to check all $\mathcal{L}+1$ feasible partitions for the pair of clusters with the minimum dissimilarity, and then choose the combination that gives the best value for the criterion function. The partition $\mathbf{C}_{\mathcal{L}}^*(\mathcal{X})$ is the result of the fusion of the chosen pair from the selected $\mathcal{L} + 1$ partition.

Finally, from Eq. (14), it is straightforward to derive a top-down greedy approximation strategy for the clustering problem by eliminating the search over the set $\Phi_{\mathcal{L}+1}$ and by checking only $\mathbf{C}_{\mathcal{L}+1}^*$, i.e.,

$$\mathbf{C}_{\mathcal{L}}^*(\mathcal{X}) = \left(\mathbf{C}_{\mathcal{L}+1}^*(\mathcal{X}) \setminus \{C'_i, C'_j\}\right) \cup \{C'_i \cup C'_j\} \tag{15}$$

where

$$(C'_i, C'_j) = \underset{\substack{C_i, C_j \in \mathbf{C}_{\mathcal{L}+1}^*(\mathcal{X}) \\ \psi(C_i, C_j) \text{ is true}}}{\operatorname{argmin}} \; d(C_i, C_j)$$

## IV. GREEDY CLUSTER MERGING FOR LOCALLY LINEAR APPROXIMATION

Our manifold approximation algorithm is based on grouping the manifold samples $\mathcal{X}$ according to their tangent spaces, in order to minimize the cost function in Eq. (10). Our new method is divided in two main steps. First, we perform the necessary preprocessing steps on the samples in order to compute the graph $G_\mathcal{X}$ and the tangent spaces $M_x$. Second, we use the graph $G_\mathcal{X}$ and the tangent spaces $M_x$'s to greedily merge the samples into feasible clusters following Eq. (15) until we reach a clustering with $\mathcal{L}$ components. The block diagram of the method is presented in Figure 2.

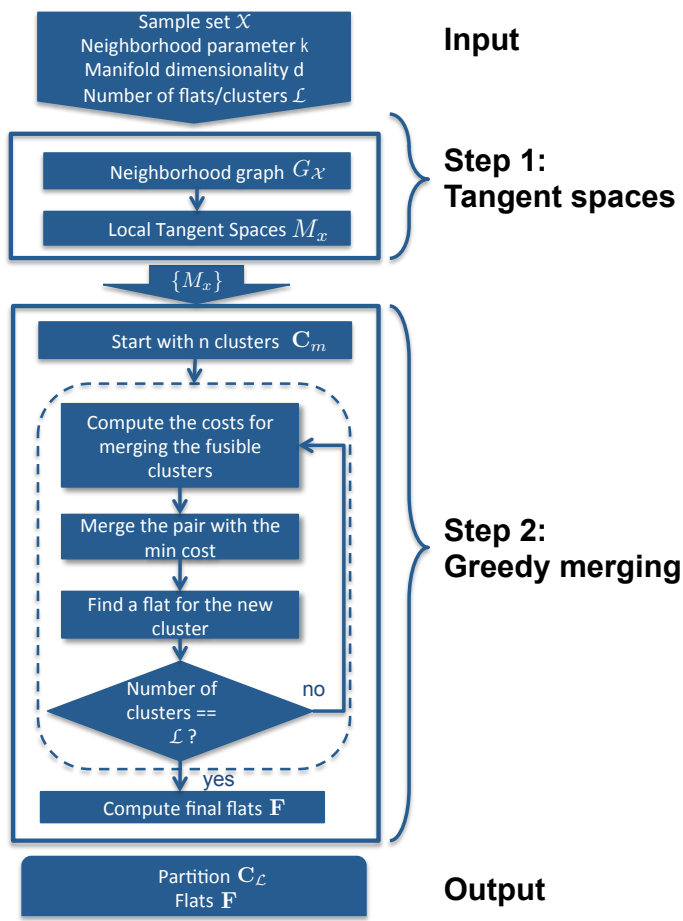

Fig. 2: The block diagram of the system.

### A. Tangent space

In the first step of the algorithm, our objective is to compute the neighborhood graph $G_\mathcal{X}$ and the tangent spaces $M_x$, one for each sample $x \in \mathcal{X}$. The neighborhood graph $G_\mathcal{X} = G(\mathcal{X}, E)$ is computed by connecting every sample $x$ to its $k$ nearest neighbors. The resulting graph $G_\mathcal{X}$ is assumed to be undirected and symmetric. For each sample $x$ we can then define a neighborhood $N_x = \{y \in \mathcal{X} : (x,y) \in E\}$ as the set of samples that are connected to $x$ by an edge in $G_\mathcal{X}$. Then, we can approximate the tangent space at $x$ by the $d$-dimensional subspace of $\mathbb{R}^N$ that best approximates the data in $N_x$. Equivalently, we compute $M_x$ as the $d$-dimensional subspace of $\mathbb{R}^N$ that best approximates the neighborhood $N_x^0$ i.e., $N_x$ shifted to the origin[1]. In other words, we need to compute the best $d$-rank approximation of the data matrix corresponding to $N_x^0$, denoted as $[N_x^0]$. Based on Eckart-Young theorem [26], this approximation is equal to the $d$-rank SVD of $[N_x^0]$. Therefore, the tangent space $M_x$ corresponds to the subspace spanned by the eigenvectors of the $d$ largest eigenvalues of $[N_x^0]$.

## B. Greedy merging

Once the graph and the tangent spaces are computed, we proceed with solving the optimization problem presented in Eq. (10). In order to minimize the cost function, we follow the method presented in Eq. (15). We start with $n = |\mathcal{X}|$ separate clusters, one for each sample. This is the optimal clustering for $n$ clusters, i.e., $\mathbf{C}_n^* = \{\{x\}, x \in \mathcal{X}\}$. Then, we reduce the number of clusters iteratively, by merging the clusters $C_i$ and $C_j$ with the minimum dissimilarity, until we reach the desired number of clusters $\mathcal{L}$.

At each iteration, there exists a set of possible mergings between the clusters in $\mathbf{C}$. The fusibility predicate defined in Eq. (2) defines the sufficient condition for a merging to be feasible: any cluster $C_i$ can be merged with any of its neighbors, i.e., the set $NG_{C_i} = \{C_j : \exists x \in C_i, \exists y \in C_j \text{ s.t } (x,y) \in E\}$. The dissimilarity between $C_i$ and $C_j \in NG_{C_i}$ is given by Eq. (13) and Eq. (4) as

$$\begin{aligned} d(C_i, C_j) &= \sum_{x \in C_i \cup C_j} D_T^2(M_x, M_{C_i \cup C_j}) - \sum_{x \in C_i} D_T^2(M_x, M_{C_i}) \\ &\quad - \sum_{x \in C_j} D_T^2(M_x, M_{C_j}) \\ &= \sum_{x \in C_i} D_T^2(M_x, M_{C_i \cup C_j}) - \sum_{x \in C_i} D_T^2(M_x, M_{C_i}) \\ &\quad + \sum_{x \in C_j} D_T^2(M_x, M_{C_i \cup C_j}) - \sum_{x \in C_j} D_T^2(T_k, M_{C_j}) \end{aligned} \quad (16)$$

---

[1] We apply a shift operator $T_{\vec{x}}$ to the whole neighborhood $N_x$, where $\vec{x}$ is the vector corresponding to the sample $x$ in $\mathbb{R}^N$. This operator moves $x$ to the origin and brings along all its neighorhood, while preserving all distances in it.

Note that since $M_{C_i}$ and $M_{C_j}$ are the mean tangents of $C_i$ and $C_j$ respectively, they minimize the sum of the square distances from the tangents in each cluster (see Eq. (6)). In other words, every other $d$-dimensional subspace would produce a higher value in the criterion function. As a result, $\sum_{x \in C_i} D_T^2(M_x, M_{C_i \cup C_j})$ is greater or equal to $\sum_{x \in C_i} D_T^2(M_x, M_{C_i})$. The same holds for the cluster $C_j$. Therefore, $d(C_i, C_j)$ is always non-negative.

However, it is costly to make all the computations of Eq. (16) for all feasible mergings. We rather use a measure that depends only on the information that is already available to the algorithm, i.e., the centers of the clusters that we have computed so far and their distances to the samples in their clusters. Moreover, since we are using a greedy top-down approach with an initial cost equal to zero, we have to ensure that, at each iteration of the algorithm, the chosen merging does only marginally increase the overall cost. Therefore, an upper bound for $d(C_i, C_j)$ that depends only on the means of the existing clusters is a suitable alternative measure for our algorithm. It contributes to reducing the complexity of the algorithm while controlling the amount of additional cost introduced at each iteration.

First, we observe that:

$$\sum_{x \in C_i} D_T^2(M_x, M_{C_i \cup C_j}) \leq \sum_{x \in C_i} D_T^2(M_x, M_{C_j}), \tag{17}$$

which means that the mean tangent of $C_i \cup C_j$ is closer to the mean tangent of $C_i$ than the mean tangent of $C_j$. This statement, which also holds if we interchange the clusters $C_i$ and $C_j$, is inevitably true. Indeed, by contradiction, if $\sum_{x \in C_i} D_T^2(M_x, M_{C_i \cup C_j})$ is larger than $\sum_{x \in C_i} D_T^2(M_x, M_{C_j})$, then $\sum_{x \in C_i \cup C_j} D_T^2(M_x, M_{C_i \cup C_j})$ is also strictly larger than $\sum_{x \in C_i \cup C_j} D_{pF}^2(M_x, M_{C_j})$. But, this contradicts the optimal character of $M_{C_i \cup C_j}$ for representing $C_i \cup C_j$ in terms of the projection distance.

Then, by substituting Eq. (17), and its equivalent form for $C_j$ in Eq. (16), we have:

$$d(C_i, C_j) \leq \sum_{x \in C_i} \left[ D_T^2(M_x, M_{C_j}) - D_T^2(M_x, M_{C_i}) \right]$$
$$+ \sum_{x \in C_j} \left[ D_T^2(M_x, M_{C_i}) - D_T^2(M_x, M_{C_j}) \right] \tag{18}$$

Moreover, by the triangle inequality:

$$D_T(M_x, M_{C_i}) \leq D_T(M_x, M_{C_j}) + D_T(M_{C_i}, M_{C_j}), \qquad \forall x \in \mathcal{X} \tag{19}$$

$$D_T(M_x, M_{C_j}) \leq D_T(M_x, M_{C_i}) + D_T(M_{C_i}, M_{C_j}), \qquad \forall x \in \mathcal{X} \tag{20}$$

Taking the square of these inequalities and summing over $C_j$ and $C_i$ respectively we get:

$$\sum_{x \in C_j} \left[ D_T^2(M_x, M_{C_i}) - D_T^2(M_x, M_{C_j}) \right] \leq 2 D_T(M_{C_i}, M_{C_j}) \sum_{x \in C_j} D_T(M_x, M_{C_j}) + |C_j| D_T^2(M_{C_i}, M_{C_j})$$

$$\sum_{x \in C_i} \left[ D_T^2(M_x, M_{C_j}) - D_T^2(M_x, M_{C_i}) \right] \leq 2 D_T(M_{C_i}, M_{C_j}) \sum_{x \in C_i} D_T(M_x, M_{C_i}) + |C_i| D_T^2(M_{C_i}, M_{C_j})$$

(21)

Substituting Eq. (21) in Eq. (18) we finally have the following dissimilarity measure:

$$d(C_i, C_j) \leq (|C_i| + |C_j|) D_T^2(M_{C_i}, M_{C_j}) \qquad (22)$$

$$+ 2 D_T(M_{C_i}, M_{C_j}) \left[ \sum_{x \in C_i} D_T(M_x, M_{C_i}) + \sum_{x \in C_j} D_T(M_x, M_{C_j}) \right],$$

which depends only on pre-computed information. By comparing Eq. (22) with Eq. (16), we can observe that Eq. (22) is indeed more computationally efficient as it involves only the means of the existing clusters and not those of the clusters after merging the fusible pairs. In our algorithm, the costs for all possible mergings at each iteration are thus computed according to Eq. (22). The clusters with the minimum estimated merging cost are then combined and the mean of the newly formed cluster is computed as shown in Section III-B. The procedure is then repeated until we reach the desired number of clusters $\mathcal{L}$.

At the end, each cluster represents a group of samples that can be well approximated by a $d$-dimensional flat. We compute the final flats for each cluster and we use the subspace spanned by the eigenvectors corresponding to the $d$ largest eigenvalues of each cluster's data matrix as representative subspace. The overall manifold approximation algorithm is summarized in Algorithm 1.

*C. Computational complexity*

We analyze here briefly the complexity of our approximation algorithm. Computing the cost of a possible merging with Eq. (22) requires only the computation of one additional tangent distance at each step. Denoting by $K_{n-\lambda}$ the number of possible mergings in the clustering $\mathbf{C}_{n-\lambda}$, the complexity of one step of the greedy merging (line 9 in Algorithm 1) requires $n + K_{n-\lambda}$ computations of tangent distances, where $n = |\mathcal{X}|$ is the initial number of clusters. Therefore, the greedy merging (lines 8-12 in Algorithm 1) will be performed in $O\left(\sum_{\lambda=1}^{n-\mathcal{L}}(n + K_{n-\lambda})\right)$ time.

We now estimate the number of possible mergings $K_{n-\lambda}$. Since, at each step of the algorithm, we perform one merging operation, we will have exactly $n - \lambda$ clusters at step $\lambda$. Moreover, each $C_i \in \mathbf{C}_{n-\lambda}$ can have a maximum size of $\lambda + 1$, and therefore we have that $C_i$ has at most $k(\lambda + 1)$ different neighbors.

**Algorithm 1** Agglomerative clustering based on differences of tangents (ACDT)

**Input:** $\mathcal{X}, k, \mathcal{L}, d$

    **Step 1**      $*$ *Preprocessing* $*$

1: Construct $G(\mathcal{X}, E)$ by connecting each element in $\mathcal{X}$ with its $k$-nearest neighbors.

2: **for all** $x \in \mathcal{X}$ **do**

3:     $N_x = \{y \in \mathcal{X} : (x, y) \in E\}$     $*$ *Compute neighborhoods* $*$

4:     $[N_x^0] = USV^T$

    where $[N_x^0]$ is the data matrix formed by the elements in $N_x$ shifted to the origin of $\mathbb{R}^N$ and $U, S, V$ are the results of its d-rank SVD.

5:     $M_x = U$     $*$ *Compute tangent spaces* $*$

6: **end for**

    **Step 2**      $*$ *Greedy computation of partition* $\mathbf{C}_\mathcal{L}^*$ $*$

7: $n = |\mathcal{X}|,\ \lambda = 0,\ \mathbf{C}_n^* = \{\{x\} : x \in \mathcal{X}\}$     $*$ *Initialization* $*$

8: **for** $\lambda < n - \mathcal{L}$ **do**     $*$ *Greedy merging* $*$

9:     $(C_i', C_j') = \underset{\substack{C_i, C_j \in \mathbf{C}_{n-\lambda}^* \\ \psi(C_i, C_j) \text{ is true}}}{\operatorname{argmin}} d(C_i, C_j)$

10:     $\mathbf{C}_{n-\lambda+1}^* = (\mathbf{C}_{n-\lambda}^* \setminus \{C_i', C_j'\}) \cup \{C_i' \cup C_j'\}$

11:     $\lambda = \lambda + 1$

12: **end for**

13: **for** $C_i \in \mathbf{C}_\mathcal{L}^*$ **do**     $*$ *Compute the final flats* $F_i$ $*$

14:     $[C_{m_i}^0] = USV^T$

    where $m_i$ is the sample mean of $C_i$, $[C_{m_i}^0]$ is the data matrix formed by the samples in $C_i$ shifted by $m_i$ and $U, S, V$ are the results of its d-rank SVD.

15:     $F_i = U$

16: **end for**

**Output:** $\mathbf{C}_\mathcal{L}^*, \mathbf{F}$

November 9, 2012      DRAFT

Thus, the number of possible mergings is at most $K_\lambda \leq \frac{1}{2}k(\lambda+1)|\mathbf{C}_{n-\lambda}| = \frac{1}{2}k(\lambda+1)(n-\lambda)$. That means that the running time $T(n)$ of the algorithm is of the following order:

$$T(n) = \sum_{\lambda=1}^{n-\mathcal{L}} O\left(n + \frac{1}{2}k(\lambda+1)(n-\lambda)\right) = O\left(\sum_{\lambda=1}^{n-\mathcal{L}}\left[n + \frac{1}{2}k(\lambda+1)(n-\lambda)\right]\right)$$
$$= O\left(\sum_{i=1}^{n-\mathcal{L}}\lambda(n-\lambda)\right) = O\left(n^3\right) \quad (23)$$

In fact, this is a rather loose bound. If one of the clusters $C_i \in \mathbf{C}_{n-\lambda}$ has a size $\lambda$, then all the other clusters have a size of 1. Therefore, it is better to use the average size of a cluster when computing the number of different neighbors, which is equal to $|\tilde{C}_i| = \frac{n}{|\mathbf{C}_\lambda|}$. Then, $K_\lambda \leq \frac{1}{2}|\mathbf{C}_\lambda|k\frac{n}{|\mathbf{C}_\lambda|} = \frac{1}{2}kn$. Therefore, on average, the running time of the algorithm is equal to $T(n) = O\left(n^2\right)$.

In comparison, if we were using the exact formula for the dissimilarity measure in Eq. (16), we would need to do one additional SVD computation and $|C_i \cup C_j|$ additional distance computations for each possible merging in $\mathbf{C}_\lambda$. Even if we assume that the computational cost of the SVD decomposition is equal or equivalent to that of a tangent distance computation, one step of the merging algorithm would require $n + (1 + |C_i \cup C_j|)K_\lambda$ new computations. Using the average estimate for $|C_i \cup C_j| = \frac{n}{|\mathbf{C}_{n-\lambda}|} = \frac{n}{n-\lambda}$, we can estimate the running time for the greedy merging to be $O\left(\sum_{\lambda=1}^{n-\mathcal{L}}(n + (1 + \frac{n}{n-\lambda})\frac{1}{2}kn)\right)$. This means that the average running time of the algorithm with the exact dissimilarity measure is $T_{SVD}(n) = O\left(\frac{1}{2}kn^2 \sum_{\lambda=1}^{n-\mathcal{L}} \frac{1}{n-\lambda}\right) = O\left(\frac{1}{2}kn^2(H_{n-1} - H_{\mathcal{L}-1})\right) = O\left(n^2 \ln n\right)$ which is higher than the average running time of our approximate algorithm[1].

## V. EXPERIMENTAL RESULTS

We have conducted two different sets of experiments to study the performance of our manifold approximation scheme. In the first one, we have tested the performance in approximating the manifold data for both synthetic and real datasets. In the second one, we have studied the use of flats for handwritten digit classification in a simple distance-based classification scheme with the MNIST dataset [32].

### A. Manifold Data approximation

We compare our scheme with two other manifold approximation approaches from the literature, namely the Hierarchical Divisive Clustering (HDC) [3] and the Hierarchical Agglomerative Clustering (HAC) [19]. The HDC algorithm starts with considering all the data as one cluster and then hierarchically

---

[1] $H_n, H_\mathcal{L}$ are the harmonic numbers of order $n$ and $\mathcal{L}$ respectively

partitions them by dividing highly non-linear clusters. As a linearity measure, it uses the deviation between the Euclidean and geodesic distances, i.e., each cluster gets a nonlinearity score that is equal to the average ratio of geodesic over Euclidean distances for all the pairs of samples in the cluster. The process continues until all existing clusters have a nonlinearity score that is lower than a given threshold. On the other hand, HAC is a bottom-up algorithm, i.e., each sample is considered at the beginning as a separate cluster and then clusters are merged iteratively until their number reduces to the desired target. At each iteration of the algorithm, the pair of clusters with the minimum distance is merged. The distance between two clusters is measured as the average geodesic distance between the samples of the one cluster and the samples of the other. Our scheme follows also a bottom-up strategy; however our distance measure is completely different than the one in [19]. The results of our tests for all three algorithms are following.

*1) Synthetic Data:* Firstly, we test the performance of our scheme in approximating synthetic manifolds. We use the Swiss roll and the S-curve dataset. The training set for both cases consists of 5000 points, randomly sampled from the manifolds. The neighborhood size $k$ is set equal to $15$ in the experiments. It is preferable to use low values for $k$, varying from $0.5\%$ to $2\%$ of the total number of samples, in order to avoid "short-circuit" effects that distort the manifold structure. In order to quantify the performance, we use the mean squared reconstruction error (MSRE) defined as $MSRE = \frac{1}{N} \sum_{i=1}^{N} ||x_i - \hat{x}_i||^2$ where $x_i$ and $\hat{x}_i$ is respectively a sample and its projection on the corresponding approximating flat, and $N$ is the total number of signals.

The MSRE versus the number of flats, for our synthetic manifolds, is presented in Figure 3. The results are averaged over 10 randomly chosen training sets. From Figure 3, we can see that our scheme approximates better the manifold structure than the other approaches. The approximation performance is better even for a small number of flats but the differences are more evident in the mid-range cases where the number of flats is between15 and 30. For higher number of flats, the difference decreases and stabilizes around 50 to 60 flats when the MSREs of the algorithms converge. The effectiveness of our method is mainly due to the use of the difference of tangents for measuring the linearity of sample sets instead of the geodesic-based criteria used by other algorithms [3], [19]. Finally, an example of the final groups computed by our algorithm is shown in Figure 4 for the case of 12 flats. In this figure, we see that the structure of the manifold is correctly preserved by the proposed manifold approximation algorithm.

*2) Natural patches:* We have also tested the performance of our scheme in approximating natural image patches since they may belong to a manifold in some applications [33]. The manifold samples are taken from the training set of the Berkeley Segmentation Dataset (BSDS) [34]. Each patch is of size $8 \times 8$ and it captures a square region of a natural image. Before approximating the manifold, we preprocess the patches so that they have zero mean and unit variance. For constructing the manifold we use 10,000 patches and k is set equal to 100.

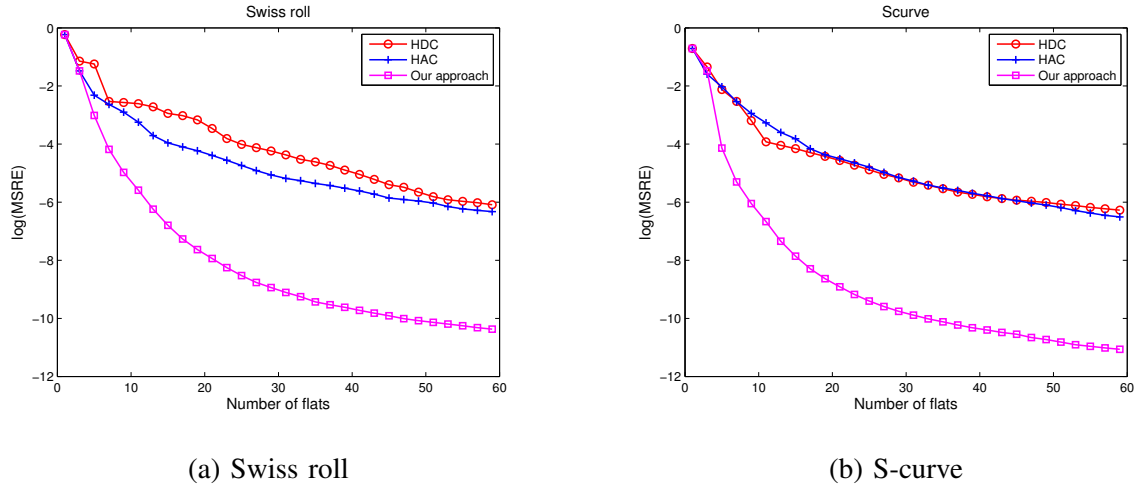

(a) Swiss roll

(b) S-curve

Fig. 3: Mean squared reconstruction error (MSRE) versus the number of flats. The error on the y-axis is shown in logarithmic scale.

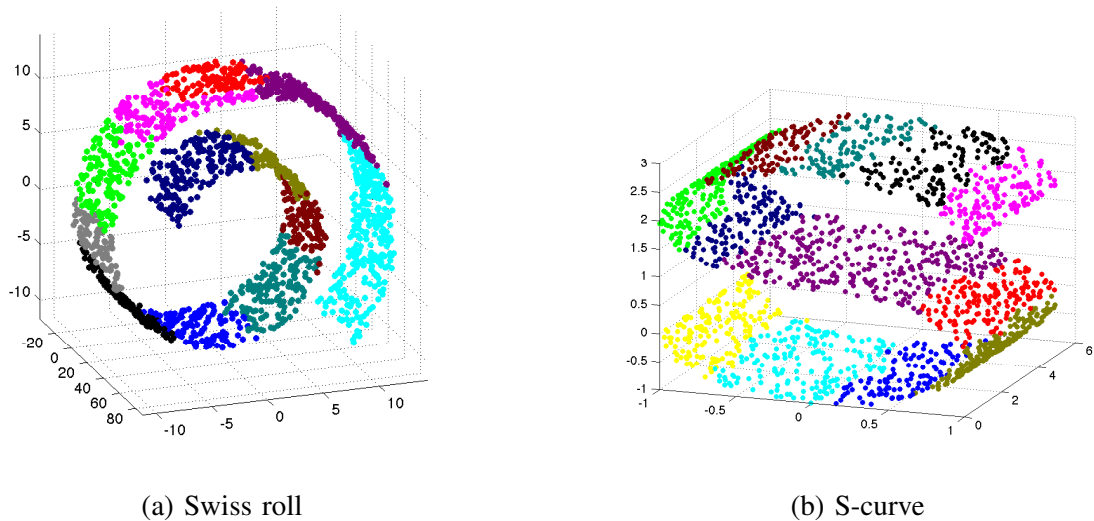

(a) Swiss roll

(b) S-curve

Fig. 4: The final groups formed by the proposed approximation algorithm with 12 flats. Each represents a different cluster of points.

The approximation performance (in terms of the MSRE) versus the number of flats is presented in Figure 5. We have plotted the approximation error for three different choices of the flats' dimensionality, i.e., $d = 16, 32$ and $60$ respectively. We can see that, in all cases, our scheme approximates significantly better the manifold structure than the other approaches. This time, the performance is higher for the whole range of the number of flats. As the number of flats approaches 400, the performance of the algorithms stabilizes especially for the case of $d = 32$ and $60$ dimensions.

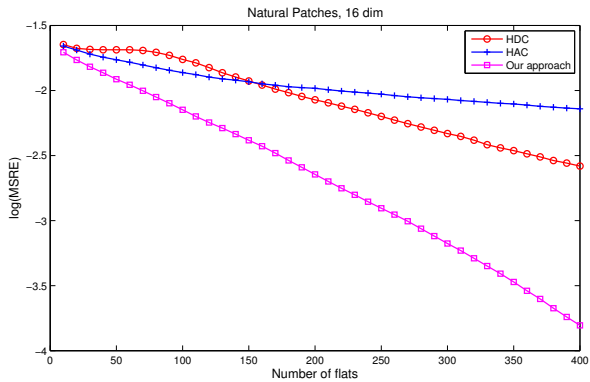

(a) 16-dimensional flats

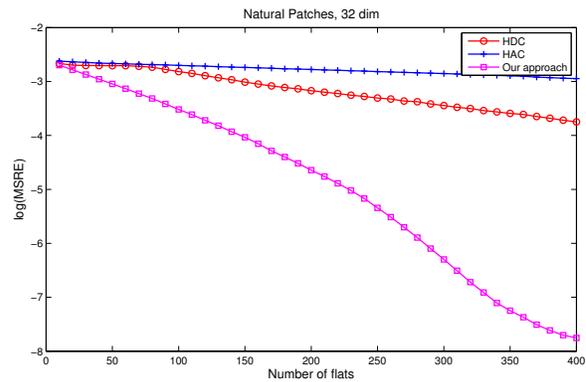

(b) 32-dimensional flats

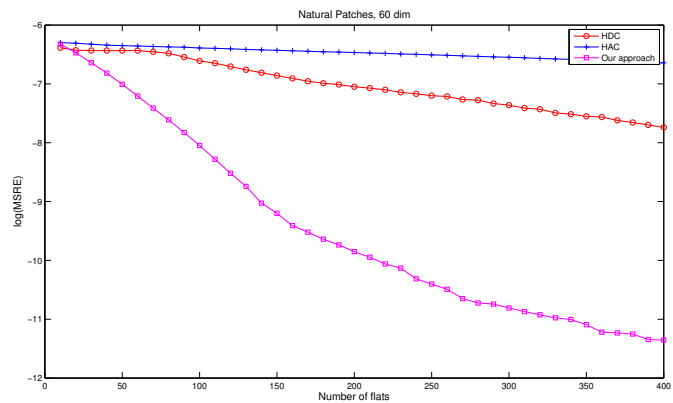

(c) 60-dimensional flats

Fig. 5: MSRE for natural patches for different choices of the flats' dimensionality. The error on the y-axis is shown in logarithmic scale.

## B. Classification

We finally check the application of our flat-based approximation to classification problems. We have built a simple classification scheme. Assuming that we have $m$ datasets, each belonging to a different class, we run at first our scheme for approximating the underlying manifold of each class with $n$ flats. We denote the set of resulting flats by $S$. Then, for each sample, we create a $m \times n$ dimensional vector of features, where each entry corresponds to the distance of a sample to a flat in $S$. Each new unclassified sample can now be projected to this flat-based feature space by the same procedure. Finally, we perform classification based on nearest neighbor criteria in this flat-based feature space.

| Feature space | Original space | PCA space |
|---|---|---|
| 95.2 % | 96.8 % | 92 % |

TABLE I: Classification performance for MNIST dataset with 10 flats per manifold.

We have checked the performance of classification for the MNIST dataset. For each digit we use 2000 random samples to construct the manifold and $n = 10$ flats to approximate it. The number of neighbors $k$ is set equal to 20 and the dimensionality of each flat is $d = 4$. For the testing we use 1000 new samples from each class. We have found that the flat-based features yield a classification rate that is comparable to that of the nearest-neighbor classification in the original space and better than the rate achieved by PCA. The results are shown in Table I. Therefore, we can say that the flats uncovered by our manifold approximation scheme manage to capture and preserve the crucial characteristics of the manifolds that could be used to discriminate samples in a space of reduced dimensionality (in our case, a space of 100 instead of 784). Finally, there is certainly a lot of space for improvement for such kind of applications by explicitly enhancing the discriminative power of the flat-based representation during the manifold approximation step.

## VI. Conclusion

We have presented a new greedy algorithm for approximating a manifold with low dimensional flats based on the difference of tangent spaces. Our method is shown to be quite powerful for manifold approximation where it outperforms state-of-the-art manifold approximation approaches. The final low-dimensional representation of signals from the manifold can be used for data compression or signal classification. In the future, we will explore ways to uncover manifold approximations that are especially useful for classification. We will also extend our method to other problems like image denoising and restoration, manifold to manifold distance computations as well as geodesic distance computations on manifolds.